\documentclass[runningheads]{llncs}
\usepackage[T1]{fontenc}
\usepackage{graphicx}
\usepackage{amsmath}

\begin{document}
\title{Stochastic differential equations for limiting description of UCB rule for Gaussian multi-armed bandits\thanks{The reported study was funded by RFBR, project number 20-01-00062.}}
\author{Sergey Garbar\inst{1}\orcidID{0000-0002-5205-5252}}
\authorrunning{S. Garbar}
\institute{Yaroslav-the-Wise Novgorod State University, Velikiy Novgorod 173003, Russia
\email{Sergey.Garbar@novsu.ru}}
\maketitle              % typeset the header of the contribution
\begin{abstract}
We consider the upper confidence bound strategy for Gaussian multi-armed bandits with known control horizon sizes $N$ and build its limiting description with a system of stochastic differential equations and ordinary differential equations. Rewards for the arms are assumed to have unknown expected values and known variances. A set of Monte-Carlo simulations was performed for the case of close distributions of rewards, when mean rewards differ by the magnitude of order $N^{-1/2}$, as it yields the highest normalized regret, to verify the validity of the obtained description. The minimal size of the control horizon when the normalized regret is not noticeably larger than maximum possible was estimated.

\end{abstract}

\section{Introduction}

We consider a multi-armed bandit (MAB) problem. MAB can be viewed as a slot machine with $J$ arms. Each one of the arms can be selected for play, which yields some random income (reward). It is suggested that a number of times that a gambler can play is known beforehand. This number $N$ is called the control horizon size. Formally MAB is a controlled stochastic process $\xi(n),n=1,2,\ldots,N$. Value $\xi(n)$ only depends on the chosen arm. Gambler’s task is to maximize the total cumulative expected reward during the time of control.

MABs and corresponding algorithms are thoroughly analyzed in \cite{lattimore}. This is a reinforcement learning problem that exemplifies the exploration–exploitation tradeoff dilemma, so it is also studied in machine learning \cite{auer}\cite{lugosi}. This problem is also known as the problem of adaptive control in a random environment \cite{sragovich} and the problem of expedient behavior \cite{tsetlin}. MABs have also been used to model problems such as managing research projects in a large organization like a science foundation or a pharmaceutical company \cite{gittins}\cite{berry}.

Further we consider a Gaussian MAB, when rewards are normally distributed with probability density functions
$$f_l\left(x\right)=\left(2\pi D_l\right)^{-1/2}e^{-\dfrac{\left(x-m_l\right)^2}{2D_l}},\ l=1,2,\ldots,J.$$

We assume that mean rewards $m_1,m_2,\ldots,m_J$ are unknown. In \cite{kolnogorov2018,garbar2021} it is shown that incorrect values for variances only slightly affect overall rewards, therefore they can be estimated during the initial steps of control. That is the reason we assume that the variances of rewards $D_1,D_2,\ldots,D_J$ are assumed to be known. Therefore, the considered MAB can be described with vector parameter $\theta=(m_1,\ldots,m_J )$.

Gaussian MAB can be useful in a batch processing setting \cite{perchet,gao,kolnogorov2012}. In that scenario gambler can only change the choice of the arm after they have used it for a given number of times (a batch). According to the central limit theorem, the sum of rewards for a batch will have a normal distribution (if batch size is relatively large) for a wide range of distributions of single-step rewards (if variances of rewards are finite). Batch processing can be also performed in parallel.

When some control strategy $\sigma$ is used, the expected cumulative reward will be lower than the maximally possible. That happens due to the fact that some number of steps will be spent to explore the distributions of arms’ rewards. Maximal cumulative reward would be obtained if the most lucrative arm was known and used on every step of the control. The expected difference between the maximal possible and the obtained rewards is called the regret and can be expressed as 
$$L_N (\sigma,\theta)=E_{\sigma,\theta} \left(\sum_{n=1}^N\left(\max(m_1,\ldots,m_J )-\xi(n) \right) \right).$$

$E_{\sigma,\theta}(\cdot)$ denotes the expected value calculated with respect to measure generated by strategy $\sigma$ and parameter $\theta$. Therefore, the goal of control is to minimize the regret.

Regret is dependent on the horizon size. To compare the strategies for different horizon sizes it is reasonable to consider the scaled regret
$$\hat{L}_N (\sigma,\theta)=(DN)^{-1/2} L_N (\sigma,\theta),$$
for $D=\max_J{D_J}$.

Single-step income on step $n$ can be expressed as 
$$\xi(n)=m_l+\sqrt{D_l} \eta_{l,n},$$
where $\eta_{l,n}$ is a standard normal random variable, and  $l=1,2,\ldots,J$, $n=1,2,\ldots ,N$.

Then cumulative reward for using $l$-th arm on step $n$ can be expressed as
$$X_l (n)=n_l m_l+\sqrt{n_l D_l} \eta,$$
where $\sqrt{n_l D_l} \eta$ is a normal random variable zero mean and variance $n_l D_l$ which we obtain when add up all of the rewards for the arm, $n_l$ is the number of times the $l$-th arm was used, $l=1,2,\ldots,J$; $n_1+\ldots +n_J=n$.

UCB strategy (as described by Lai \cite{lai}) with the parameter $a$ is defined as selection of arm maximizing the value
$$U_l (n)=\dfrac{X_l(n)}{n_l} +\dfrac{\sqrt{aD_l  \log(n/n_l} )}{\sqrt{n_l }},l=1,2,\ldots,J; n=1,2,\ldots,N.$$

Strategy prescribes to use each of the arms once in the beginning of control to make the initial estimations regarding the mean rewards.

Presence of slowly growing term $\sqrt{2D_l \log(n/n_l)/n_l }$ ensures the necessity to use every arm from time to time even if the estimated mean reward $X_l (n_l )/n_l$ from its usage so far was lower than from the other arms due to some reason. That is the way the UCB strategy negotiates the exploration–exploitation dilemma: the gambler strives to maximize his or her reward by using the arm with the highest estimated mean, but also collects the information about the less explored arms.

We aim to build a limiting description (with a system of stochastic differential equations) of UCB strategy for a Gaussian multi-armed bandit when $N\to\infty$.

\section{System of stochastic differential equations to describe UCB strategy for Gaussian two-armed bandit}

First, we give a limiting description for the case of two-armed bandit.

In \cite{garbar2020novsu,garbar2020c} an invariant description of a UCB strategy for a MAB was given. Invariant description scales the control horizon to a unit size by scaling the step number to smaller values (proportional to $N$). If $N\rightarrow\infty$, then step value is infinitely small.

Differentials of cumulative reward for using $l$-th arm can be expressed as
$$dX_l=m_ldt_l+\sqrt{D_l}dW_l,\ l=1,2.$$

Here $W_1,W_2$ are independent Wiener processes, $t_l$ is a continuous variable that corresponds to the usage ratio of each arm. Each of those equations correspond to a Wiener process with a constant drift.
At each moment only one of values $t_1,t_2$ can be increased as only one of the arms is used. Therefore, we consider two areas: in the first $t_1$ increases with time $t$, in the second $t_2$ increases. We’ll use indicators $I_1\left(t\right),I_2\left(t\right)$ for convenience.

If the first arm is chosen, then 
\begin{center}
	$\begin{cases}
		dt_1=dt, \\
		dt_2=0,
	\end{cases}$ and 
	$\begin{cases} 
		I_1(t)=1, \\
		I_2(t)=0.
	\end{cases}$
\end{center}

If the second arm is chosen, then 
\begin{center}
	$\begin{cases}
		dt_1=0, \\ 
		dt_2=dt, 
	\end{cases}$ and 
	$\begin{cases}
		I_1(t)=0, \\
		I_2(t)=1. 
	\end{cases}$
\end{center}

We’ll take notice that $I_2\left(t\right)=1-I_1\left(t\right)$ and $dt_2=dt-dt_1$. Also remember that $t_2=t-t_1$: one and only one arm is chosen at every moment.

Each of the areas is determined according to the chosen strategy. UCB rule can be rewritten as 
$$U_l\left(t,t_l\right)=\dfrac{X_l\left(t_1\right)}{t_l}+\dfrac{\sqrt{aD_l\log{\dfrac{t}{t_l}}}}{\sqrt{t_l}},\ l=1,2.$$

The area where the first arm is chosen will be defined by the condition $U_1\left(t,t_1\right)>U_2\left(t,t_2\right)$. Otherwise the second arm is chosen. To write this condition formally we use the Heaviside step function $H(\cdot)$ notation. Its value is zero for negative arguments and one for positive arguments.

For the UCB rule the indicator $I_1\left(t\right)$ will be the function of arguments $t,t_1,X_1,X_2$:
$$I_1\left(t,t_1,X_1,X_2\right)=H\left(\dfrac{X_1}{t_1}+\dfrac{\sqrt{aD_1\log{\dfrac{t}{t_1}}}}{\sqrt{t_1}}-\dfrac{X_2}{t-t_1}-\dfrac{\sqrt{aD_2\log{\dfrac{t}{t-t_1}}}}{\sqrt{t-t_1}}\right).$$

Using this notation, we write the system of two Itô stochastic differential equations and one ordinary differential equation equations to describe the UCB strategy for Gaussian two-armed bandit as
$$
\begin{cases}
	dX_1=I_1m_1dt+\sqrt{D_1}dW_1, \\ 
	dX_2=\left(1-I_1 \right)m_2dt+\sqrt{D_2}dW_2, \\
	dt_1=I_1dt.
\end{cases}
$$

\section{Simulation results}

A series of Monte-Carlo simulations was performed to verify how the obtained system of equations describes the behavior of the Gaussian two-armed bandit when the UCB strategy is used. Without the loss of generality, we consider a Gaussian two-armed bandit with the zero mean reward for the first arm $m_1=0$. We observe how the normalized regret depends on the value of the mean reward for the second arm. Also, we assume $D_1=D_2=D=1$: in \cite{garbar2020c} it is shown that the maximum normalized regret is observed when the variances of the arm rewards are equal. Also, we set $a=1$.

Further we consider the case of “close” distributions of rewards as it is when the highest values of regret are observed \cite{vogel}. Its definitive feature is that the difference of mean rewards has order $N^{-1/2}$:
$$\left\{m_l=m+c_l\sqrt{D/N};m\in{R},\left|c_l\right|\le C<\infty,l=1,2\right\}.$$
Without the loss of generality we set $c_1=0$.

For the Gaussian multi-armed bandit simulations, we consider the cases of horizon sizes $N=200,\ 400$. Figure 1 contains the plots of normalized regret vs different values of $c_2$. Data for the plots are averaged over 10000 simulations. Figure 1 also shows the normalized regrets calculated for numerical solutions of presented stochastic differential equations for different values of $c_2$, averaged over 10000 simulations. We used the Euler–Maruyama method with step sizes of ${10}^{-3}$ and $5\cdot{10}^{-4}$.

\begin{figure}
	\centering
	\includegraphics[width=\textwidth]{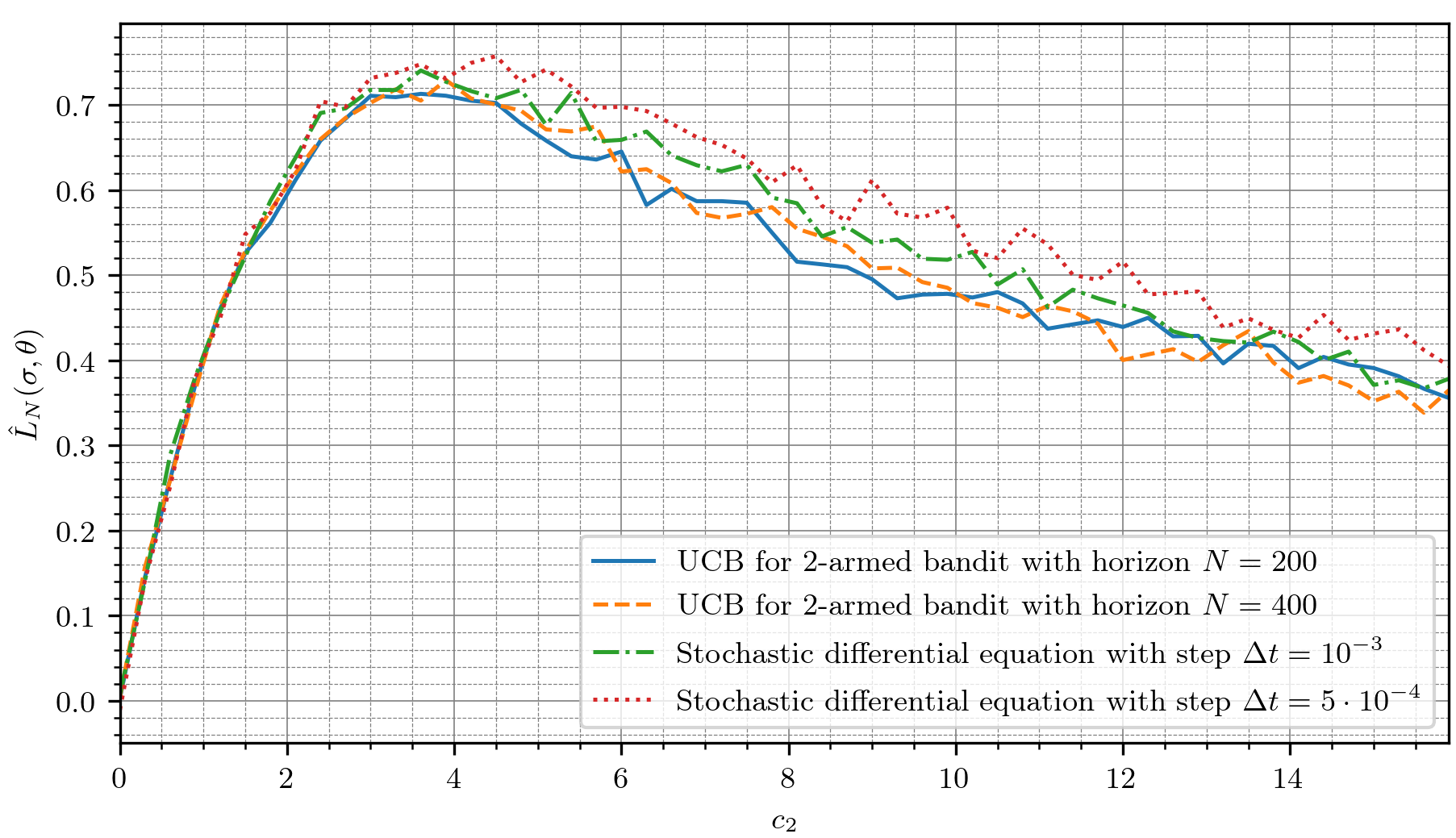}
	\caption{Normalized regret for various differences of mean rewards (parameterized by $c_2$) obtained by Monte Carlo simulations and as solution of system of stochastic differential equations} \label{fig1}
\end{figure}

We see that normalized regret is about the same for different horizon sizes. Also, the obtained system of equations gives the fitting description of the UCB strategy. We note that maximal normalized regret  (approximately 0.73) in these simulations is observed when $c_2\approx3.6$.

Having the limiting description allows to answer the question at what horizon size $N$ the normalized regret will not be noticeably larger than found maximal normalized regret. This question is of importance as the batch version of the algorithm (reported in \cite{garbar2020novsu,garbar2020c}) allows to vary the batch sizes. Parallel processing is possible in that case, and it can result in shorter processing times.

Figure 2 shows maximum normalized regret (in domain of close distributions of rewards) vs the control horizon size (left plot shows $N\in\left[3,100\right]$, right plot shows $N\in\left[36,100\right]$). Data is averaged over 10000 simulations.

\begin{figure}
	\centering
	\includegraphics[width=\textwidth]{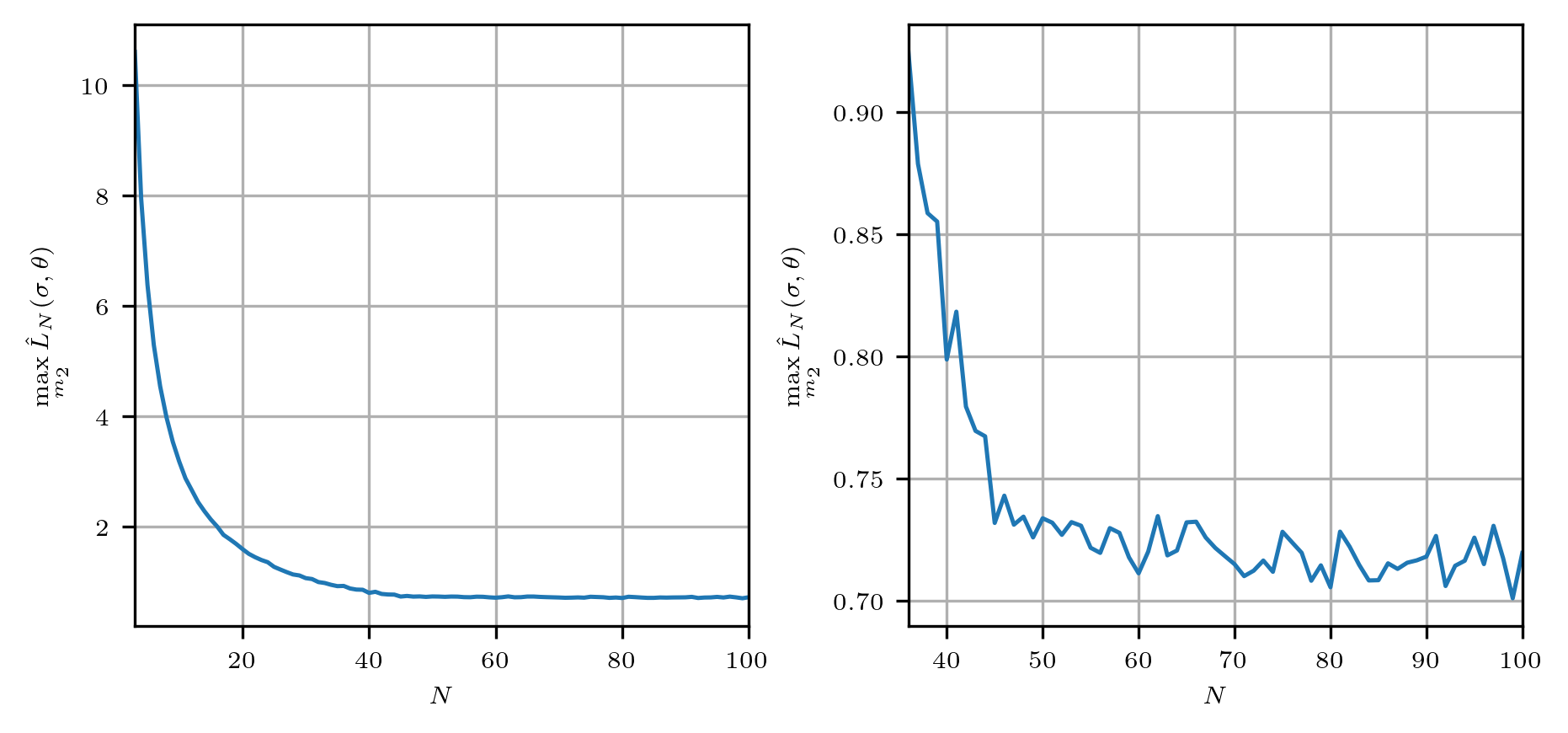}
	\caption{Maximum normalized regret vs the control horizon size $N$} \label{fig2}
\end{figure}

We see that after $N=45$ the normalized regret exceeds the value found via the limiting description by no more than $2\%$. That means that for the batch processing (including parallel), the batch size may be chosen fairly large as long as the number of processed batches stays small of moderate (i.e. 50 or more).

\section{System of stochastic differential equations to describe UCB strategy for Gaussian MAB}

Further we obtain a limiting description with a system of stochastic differential equations for the UCB strategy for Gaussian multi-armed bandit with $J$ arms using the similar reasoning as before. Considered MAB has unknown mean rewards $m_1,\ldots,m_J$ and known variances of rewards $D_1,\ldots,D_J$.

First, we express $t_J$ as 
$$t_J=t-\sum_{i=1}^{J-1}t_i.$$

Indicator for the arm will be equal to one if the corresponding upper confidence bound is greater than the confidence bounds for all other arms. We express it as the product of step functions
$$
\begin{aligned}
	I_l(t,t_1,&\ldots,t_{J-1},X_1,\ldots,X_J)= \\ 
	&=\prod_{\substack{i=1 \\ i\neq l}}^{J}{H\left(\dfrac{X_l}{t_l} +\dfrac{\sqrt{a D_l \log\dfrac{t}{t_l}}}{\sqrt{t_l}}-\dfrac{X_i}{t_i} -\dfrac{\sqrt{aD_i  \log\dfrac{t}{t_i}}}{\sqrt{t_i}}\right) },
\end{aligned}
$$
for $l=1,\ldots,J-1$.
$$I_J\left(t,t_1,\ldots,t_{J-1},X_1,\ldots,X_J\right)=1-\sum_{i=1}^{J-1}{I_i\left(t,t_1,\ldots,t_{J-1},X_1,\ldots,X_J\right)}.
$$

Then the system of $J$ stochastic differential equations and $J-1$ ordinary differential equations that describes the UCB strategy takes form 
$$
\begin{cases}
	dX_1=I_1m_1dt+\sqrt{D_1}dW_1, \\
	… \\
	dX_{J-1}=I_{J-1}m_{J-1}dt+\sqrt{D_{J-1}}dW_{J-1}, \\
	dX_J=\left(1-\sum_{i=1}^{J-1}I_i\right)m_Jdt+\sqrt{D_J}dW_J, \\
	dt_1=I_1dt, \\
	… \\
	dt_{J-1}=I_{J-1}dt.
\end{cases}
$$

Presented system of equations also can be used to determine the horizon size at which the normalized regret converges. That can be used as the number of processed batches for the batch version of strategy, described in \cite{garbar2020novsu}.

We consider a case of three-armed bandit as an example. We also consider the case of close distribution of rewards, described in section 3. The highest normalized regret is observed when $c_1=c_2=0$ and $c_3\approx4.5$ \cite{garbarkolnogorov2020asmda}. This happens as in this situation the strategy must distinguish between the best arm and two competing arms with fairly similar but lower rewards. Simulations show that the normalized regret converges at horizon size $N=70$, i.e. is no more than $2\%$ greater than its value for larger horizon sizes.

\section{Conclusion}

We built and examined a limiting description (for horizon size $N\rightarrow\infty$) with a system of stochastic differential equations for the UCB strategy for the Gaussian multi-armed bandit. 

A series of Monte-Carlo simulations was performed to verify the validity of obtained results for the case of two-armed bandit in the case of “close” distributions of rewards. Maximum normalized regret for the case $N\rightarrow\infty$ was determined. The minimal size of the control horizon when the normalized regret is not noticeably larger than maximum possible was estimated.

\subsubsection{Acknowledgements} The reported study was funded by RFBR, project number 20-01-00062.

\bibliographystyle{splncs04}
\bibliography{mybibfile}

\end{document}